\documentclass{article}

\usepackage{arxiv}

\usepackage[utf8]{inputenc} 
\usepackage[T1]{fontenc}    
\usepackage{hyperref}       
\usepackage{url}            
\usepackage{booktabs}       
\usepackage{amsfonts}       
\usepackage{nicefrac}       
\usepackage{microtype}      

\usepackage{graphicx,float}
\usepackage[toc,page]{appendix}

\usepackage{doi}
\usepackage{xcolor}

\usepackage{amsmath,amssymb}
\usepackage{multicol}
\usepackage[export]{adjustbox}
\usepackage{calc}

\usepackage[final]{changes}

\usepackage{natbib}
\title{Context-dependent communication under environmental constraints}

\author{Krzysztof Główka\thanks{The corresponding author}\\
	Human Interactivity and Language Lab\\
	Faculty of Psychology\\
	University of Warsaw \\
	Poland \\
	\texttt{krz.glowka@gmail.com}
	\And
	Julian Zubek \\
	Human Interactivity and Language Lab\\
	Faculty of Psychology\\
	University of Warsaw \\
	Poland \\
	\texttt{j.zubek@uw.edu.pl}
	\And
	Joanna Rączaszek-Leonardi \\
	Human Interactivity and Language Lab\\
	Faculty of Psychology\\
	University of Warsaw \\
	Poland \\
	\texttt{raczasze@psych.uw.edu.pl}
}

\date{}

\renewcommand{\shorttitle}{Context-dependent communication under environmental constraints}

\hypersetup{
pdftitle={Context-dependent communication under environmental constraints},
pdfauthor={Krzysztof Główka, Julian Zubek, Joanna Rączaszek-Leonardi},
}

\begin{document}
\maketitle

\begin{abstract}
There is significant evidence that real-world communication cannot be reduced to sending signals with context-independent meaning. In this work, based on a variant of the classical \citet{lewis_1969} signalling model, we explore the conditions for the emergence of context-dependent communication in \replaced{an agent-based situated model}{a situated scenario}. In particular, we demonstrate that pressure to minimise the vocabulary size is sufficient for such emergence. 
At the same time, we study the environmental conditions and cognitive capabilities that enable contextual disambiguation of symbol meanings.
We show that \added{a) regularities in the context are not necessary for context-dependent communication and that b)} environmental constraints on the receiver's referent choice can be unilaterally exploited by the sender, without disambiguation capabilities on the receiver’s end. Consistent with common assumptions, the sender’s awareness of the context appears to be required for contextual communication. We suggest that context-dependent communication is a situated multilayered phenomenon, crucially influenced by environment properties such as \replaced{distribution of objects in the context}{distribution of contexts}. The \added{computational} model developed in this work is a demonstration of how signals may be ambiguous out of context, but still allow for near-perfect communication accuracy.
\end{abstract}


\begin{multicols}{2}

\section{Introduction}

In language sciences and philosophy, there is a marked tension between the perceived invariance of the meaning of natural language and its inevitable interaction with the actual context of usage. In this vein, it is common to distinguish between semantics, concerned with the \replaced{context-independent}{standalone} meaning, and pragmatics, investigating \replaced{the contribution of the context}{situational context} (for an overview, see \citet{korta2020}). In much of linguistic work, the focus is on semantics, while pragmatics is peripheral \citep{steffensen2018}.

Although such a relatively narrow framing of natural language may have been useful for a number of research goals, there is ample evidence that, in real-world communication, \replaced{context-independent}{standalone} meaning is an exception rather than a rule. Human interlocutors seem to readily produce messages that are ambiguous \citep{wasow2005puzzle}, relying on the context to complete their meaning \citep{Barwise1983-BARSAA}. \added{The mode and extent of this dependence vary widely: the context guides our understanding of homonyms such as \textit{bat} (are we playing baseball or watching the night sky?); it allows us to interpret approximate quantities such as \textit{high price} (how much do I usually spend on this?); and deictic expressions like \textit{this} are perhaps the most vivid examples, as they cannot function without a situational or linguistic context.} \replaced{Yet}{In fact,} humans understand language in context so effortlessly that potential \replaced{misunderstandings}{ambiguities} are often only noticed by the linguist \citep{ferreira2008ambiguity}.

In this work, we focus on two questions. First, what pressure is sufficient for the emergence of context-dependent communication (why it may be preferred over unambiguous communication)? Second, what environmental conditions and cognitive capabilities enable such communication? To address these issues, we turn to agent-based models as tools to examine various factors shaping language evolution in simplified and idealised simulations \citep{cangelosi2001evolution, steels2011modeling, cangelosi2012simulating}. We adopt a variant of the \citet{lewis_1969} signalling game -- a cooperative scenario in which the sender's task is to communicate the referent that the receiver must identify in a limited set. Within this model, we introduce the communication context as a set of possible referents, selected at random in each communication episode.

Even in the simple scenario of the signalling game, context may play a crucial role in at least two aspects: (a) it offers additional information that can be exploited by the sender and the receiver, and (b) context directly constrains the receiver's action choice, which may impact the agents' communicative strategy regardless of the receiver's information-processing capabilities. We can illustrate this distinction by building upon an example given by \citet{Searle1980-SEATBO-2}: when someone hears the utterance ``we need to cut it'' while standing on a lawn, they are likely to think about a mower. The same sentence uttered during a birthday party can be interpreted as a request to fetch a knife to cut a cake. However, when the sender knows that the right cutting tool, a mower or a knife, is already present at the receiver's hand, they can be sure that the receiver will understand the utterance correctly even without knowing the context, as the possible course of action is already narrowed down.

While previous research on context-dependent communication has usually assumed agents' nontrivial internal information-processing capabilities, we focus on the enabling constraints external to the agents and carry out experiments with the receiver's minimal skills. We observe that the following conditions are sufficient for the development of context-dependent communication in this setting: (a) information on the receiver’s context is provided to the sender and (b) pressure is placed on the agents to minimise their vocabulary. Under these conditions, agents develop a language that is partly ambiguous and rely on the context for disambiguation rather than invent a fully precise language.

\section{Why does linguistic ambiguity arise?}
\label{sec:ling_context}

Linguists' focus on context-independent language is reflected and arguably sustained by the commonly adopted metaphors of communication as a transmission problem and the resulting treatment of language as a code \citep{reddy1979}. The accompanying pervasive assumption of stable correspondence between language utterances and their meaning has roots in the long history of focus on language generativity \citep{katz1963structure}. More pragmatically and ecologically-oriented works argue for a different picture, where natural language utterances are commonly ambiguous out of context -- but they are successfully understood by humans based on the shared situational context \citep{sperber1986}. As \citet{Barwise1983-BARSAA} indicated, while it is important that we can express an infinite number of propositions with a finite number of elements, due to the possibility of combination and recursion, it is equally amazing that we can express an infinite number of meanings with one expression due to its dependence on context \added{(bringing to mind examples of entirely context-dependent expressions such as \textit{this})}.

Arguably, the conventional meaning of expressions can be extended to novel contexts due to their ambiguity \citep{brochhagen2015improving}\replaced{. Recent works propose that expression meaning is negotiated in interaction \citep{chater2022grammar} and that language, rather than transmitting clear and bounded meanings, has the role of steering or controlling interactive events \citep{raczaszek2012constraints}}{; and expression meaning is created in interaction \citep{chater2022grammar}, where language has the role of steering or controlling interactive events rather than transmitting clear and bounded meanings \citep{raczaszek2012constraints}}. These directions in the development of the theory of language, from the transmission model to viewing language as a \added{prevalently ambiguous out of context,} creative\replaced{ interaction control, situated in a particular environment}{, situated interaction control}, make understanding of the nature of context and its contribution to linguistic meaning and structure an urgent task~\footnote{Paradoxically, from such a standpoint it seems more challenging to account for the partial meaning context-independence than to acknowledge the effects of situational context; for example, how it is possible that we have a sense that utterances refer to something seemingly unrelated to the current circumstances, possibly distant in time and space \citep{raczaszek2018language}.}. \added{In this work, we investigate the role of the environment and, consequently, focus on the extra-linguistic context, understood as the current or past composition of the environment.}

\added{
While the perspective adopted in this study foregrounds the positive and neutral aspects of linguistic ambiguity, we note that previous studies often associated ambiguity with miscommunication. As we see no compelling argument to make a clear distinction between these two cases, we consistently use the term \textit{ambiguity} to refer to all cases of meaning uncertainty.
}

\citet{piantadosi2012} propose that ambiguity arises under pressure for communicative efficiency. They consider the scenario of a sender and receiver in a shared context. If the context is at least slightly informative about the possible meanings of the sender's message, the sender does not need to convey full information for the receiver to disambiguate the message, that is, infer its meaning from the context. Consequently, if symbol production is costly and the receiver's inference is relatively cheap, it is suboptimal on the sender’s part to ensure that their utterance can only be interpreted in exactly one way. A shorter, partially ambiguous utterance will be preferred and will still be informative enough for the receiver. Additionally, if there are differences in the production cost of certain symbols, the sender may choose more economic, but more ambiguous ones.

While the bottleneck indicated by Piantadosi and colleagues concerned language production, \citet{kirby_2015} argue for an informational bottleneck in cultural transmission. They demonstrate that natural limitations in the cognitive abilities of language learners impose pressure for \textit{simplicity} on a language system. Thus, despite the pressure for \textit{informativeness}~\footnote{We follow the use of terms in \citet{winters2018contextual}; the original terms were \textit{compressibility} and \textit{expressivity} \citep{kirby_2015}.}, arising from the use of language, languages will never become perfectly informative about the intended meaning. Rather, they will gravitate towards a point of optimal trade-off between the competing pressures, where a degree of context dependence is to be expected. Under such a trade-off, language ambiguities can make communication less effective but still functional in general \citep{santana2014ambiguity, o2015ambiguity}.

Intuitively, if language meaning relies on the context, then the composition of contexts, in which the interlocutors find themselves, should also play a role for ambiguity. Imagine a simple interaction between a shop assistant and a customer asking for a specific product visible on the display. Here, successful communication requires \textit{discrimination} of the intended referent from other possible referents (e.g. \citet{olson1970language}). \citet{winters2018contextual} demonstrated that in this kind of situated setting, more predictable contexts allow for minimising information transmission through linguistic means. They manipulated the sender's access to context and contextual object distribution in artificial language experiments, where pairs of participants were tasked to communicate about the target referent. When the possible referents in the context were random, the emergent signalling system was highly \replaced{context-independent}{autonomous}. However, when objects within each context were always of the same colour and the sender could see the receiver's context, the emergent languages became ambiguous, not encoding colour information, and thereby relying on the stable object distribution in the context.

In summary, several factors seem to crucially contribute to the emergence of ambiguous language. One of them is the situational context shared between interlocutors, which is a source of additional information independent of what is conveyed by language. The second is an additional pressure that makes the language not only informative but also reasonably simple. The last is the regularity of the environment, which determines the relevant distinctions to be made by language.

\section{Context in signalling games}
\label{sec:context_related_work}

The models relevant for this work follow the general schema of the influential \citet{lewis_1969} signalling game, where the sender and receiver learn to communicate about objects in their environment through rewards for communication success.
In the original formulation of the game, as well as in its many extensions, there is no shared context between the agents. Under such circumstances, the only protocols that guarantee communication success are those that unequivocally map objects to signals, and they predictably emerge in simulations of evolutionary dynamics with realistic perturbations \citep{skyrms_2010, santana2014ambiguity}.

In some more recent models of visually grounded language emergence, neural agents faced an image discrimination task with shared context that directly constrained the referent choice (e.g. \citet{lazaridou2017, lazaridou2016towards, lazaridou_2018, havrylov2017emergence, bouchacourt2018agents}). The size of the context ranged from just two objects (e.g. \citet{lazaridou2016towards}) to $128$ \citep{havrylov2017emergence}. In these works, due to their engineering motivation, the only optimisation goal for agent policies was communication success. The results of \citet{guo2021expressivity} suggest that the constrained choice of referents can itself lead to the emergence of context-dependent languages. Nevertheless, context dependence was not controlled for nor reported in these works.

Some works in evolutionary game theory investigated the origins of language ambiguity in signalling games where additional costs were included in the reward function. This included an additional cost imposed on the sender for using a larger vocabulary \citep{santana2014ambiguity, muhlenbernd2021evolutionary} and / or the cost related to disambiguation imposed on the receiver \citep{muhlenbernd2021evolutionary}. The communicating agents were provided with shared contextual information. The results show that partial ambiguity is evolutionarily the most likely strategy when the sender's signalling cost is higher or comparable to the receiver's disambiguation cost. In a similar vein, \citet{brochhagen2020signalling} showed that the context can be exploited by the sender to use its preferred but more ambiguous forms. In these game-theoretic studies, the receiver had the ability to infer the intended meaning of the sender, in congruence with \citet{piantadosi2012}. However, the direct impact of the environment on context-dependent communication was not accounted for because of the unlimited referent choice.

While the receiver's inference is undoubtedly important for context-dependent communication \citep{frank2012}, direct environmental constraint on the choice of possible referents substantially modifies the task \citep{winters2018contextual, guo2021expressivity}. In a situated communicative setting with a constrained referent choice, the speaker's signal needs to be only as precise as to discriminate the referent from other present objects, and the receiver's task is arguably less challenging, as the agent's inference is partly ``offloaded'' to the environment. Therefore, to investigate the role of the environment for context dependence, we use a model in which agents have minimal skills and in which the pressure for vocabulary size and sender's access to context will be manipulated.

\added{
Importantly, we use an entirely random distribution for objects in the context. This is for two reasons. Firstly, it is to be expected that regularities in the encountered contexts can yield more ambiguous but still effective signals \citep{guo2021expressivity} but the possible outcomes for an entirely random object distribution under vocabulary size pressure are less clear. Secondly, we aimed for the most general conditions for the ambiguous communication to emerge and thus made as few assumptions as possible.
}

\section{Contextual Lewis signalling game}
\label{sec:modelling_context}

Let us define a contextual variant of the probabilistic Lewis signalling model. There are two agents: sender and receiver. The sender perceives one of the alternative states of the world $s \in S$, and emits one of the available signals $\sigma \in \Sigma$ with probability $f(\sigma | s)$. The receiver chooses a response $r \in R$, based on the received signal, with probability $g(r | \sigma)$. Upon receiving a correct response $r$ for a given state $s$ both agents are rewarded. Agent policies are represented through functions $f$ and $g$ and can be modelled, for example, using neural networks. Through adaptive learning, agents can reach an equilibrium in which signals are used for functional communication. Signals are conventional in the sense that $s$--$\sigma$ and $\sigma$--$r$ mappings are arbitrary but must be consistent. We can say that the signalling protocol established in a Lewis signalling game is unambiguous if there is a unique mapping from signals $\sigma$ to states $s$. Such a setting and its many varieties have been a kind of general testing ground for investigating various aspects of language emergence.

\replaced{
Let us call a non-contextual variant of this game where the receiver's output belongs to the same space as the sender's input (namely, the response belongs to the set of possible world states, $r \in S$), a \textit{reconstruction game}, following recent usage of the term by \citet{guo2021expressivity}.
In the reconstruction game, the receiver has to learn to reconstruct the target solely from the signal emitted by the sender, without any contextual cues (e.g. \citet{chaabouni_2020, kharitonov2020emergent}).
}
{
Within recent neural models, a game that follows this schema without extending it with contextual elements can be called \textit{reconstruction game}, following \citet{guo2021expressivity}.
In these works, the receiver's output belongs to the same space as the sender's input, namely, the response belongs to the set of possible world states: $r \in S$ (e.g. \citet{chaabouni_2020, kharitonov2020emergent}).
In the reconstruction game, the receiver has to reconstruct the target solely based on the received signal -- there are no contextual cues of any kind. 
}

In contrast, in another popular scenario, which we call \textit{discrimination game}~\footnote{\citet{guo2021expressivity} have called this scenario simply \textit{referential game}. However, this name has been used across a wide variety of contexts and fields. To avoid confusion and \replaced{distinguish this setting from reconstruction game more accurately}{capture the essence of those scenarios}, we choose the term \textit{discrimination}.} the receiver's task is to identify the target from among the provided set of objects which constitutes the context. We will use the word \textit{context} in this narrow technical sense. The world state is multi-object, that is, if $O$ is the set of all possible objects, the world state consists of the context $C \subset O$ and the target $t \in C$ -- what is communicated is only a part of the world state.

Within a discrimination game, we can consider a receiver that is not ``aware'' of the context, but is still constrained by it, since only some objects can be selected. For such a case, we need to define an auxiliary selection function:
\[
\iota_C(\mathbf{z})_i = \begin{cases}
        \dfrac{e^{z_i}}{\sum_{r_j \in C} e^{z_j}} & \text{if } r_i \in C\\
        0 & \text{if } r_i \not\in C
    \end{cases}
\]
The function $\iota_C$ operates on a vector of probabilities $\mathbf{z}$ corresponding to different responses $r \in O$. It selects only valid responses that belong to a given context and normalizes the probabilities.

On a spectrum of perceptual availability of the context, we can distinguish the following variants of the sender--receiver pairs playing a discrimination game, from the least to the most contextual:

\begin{enumerate}
    \item Neither sender nor receiver have access to the context. Sender's signal is generated with probability $f(\sigma|t)$, receiver's response is generated with probability $\iota_C(g(r|\sigma))$.
    \item Sender has access to the context, but the receiver does not. Sender's signal is generated with probability $f(\sigma|t, C)$, receiver's response is generated with probability $\iota_C(g(r|\sigma))$.
    \item Both sender and receiver have access to the context. Sender's signal is generated with probability $f(\sigma|t, C)$, receiver's response is generated with probability $g(r|\sigma, C)$.
\end{enumerate}

Under variant 1) agents cannot use contextual information at all. The optimal solution is to use unambiguous signalling, where every world state is represented by a unique signal. Under variant 3) sender and receiver share context, and may use it as a common ground. This, in principle, opens a possibility for ambiguous, yet efficient, communication protocols, such that the sender emits the signal conditional on the context and the receiver disambiguates the signal in a process of inference within the particular context \citep{piantadosi2012}.

Variant 2) is an interesting middle ground between these two possibilities. Since sender knows which objects are available for the receiver to choose from, it may safely use an ambiguous signal when other probable referents of the signal are not present in the current context. Signalling protocol that is highly ambiguous in the entire object space may still be optimal for the vast majority of contexts. Thus, accurate context-dependent communication may emerge even without inferential capabilities on the receiver's side. This is the variant that we implement in our simulations.

\section{Model architecture}
\label{sec:contextual_model}

Agents are modelled as multi-layer neural networks. We consider two agent pairs, corresponding to variant 1 and 2 from section \ref{sec:modelling_context}, including the same type of receiver but differing with respect to the sender's access to the context beyond the target object. Importantly, the receiver is designed to be unable to infer signal meaning in context.

\begin{figure}[H]
    \centering
    \begin{tabular}{cc}
        (a) \raisebox{1ex-\height}{\includegraphics[scale=0.5]{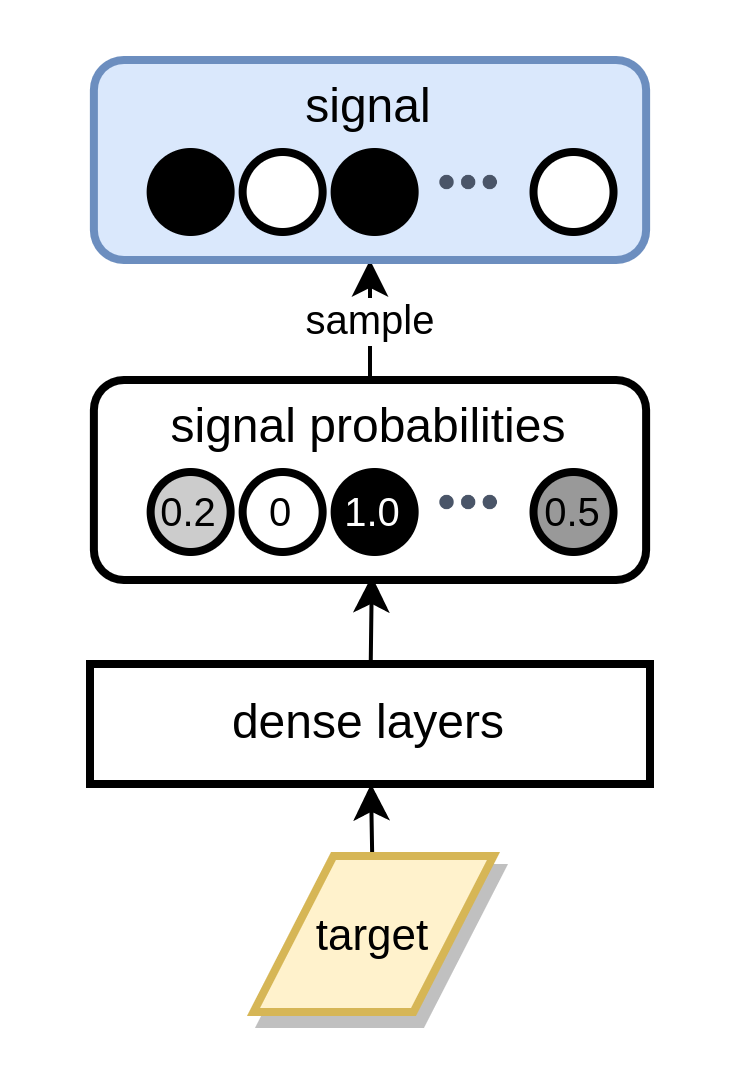}}
        & (b) \quad \raisebox{1ex-\height}{\includegraphics[scale=0.37]{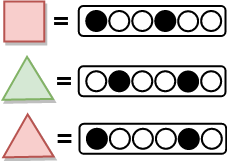}}
    \end{tabular}
    \caption{\textbf{The target-only (T) sender} (a) and example vector encodings of object properties (b).}
    \label{sender_t}
\end{figure}

\textbf{Target-only (T) sender}.  A simple feed-forward network of fully-connected layers, which receives only the target as the input and maps it to probabilities of activation of each symbol (Fig.~\ref{sender_t}). Since this sender cannot access the context, it shares no common information with the receiver, and thus it is not expected to be able to communicate in a context-dependent manner in any setting.

\begin{figure}[H]
    \centering
    \includegraphics[scale=0.5]{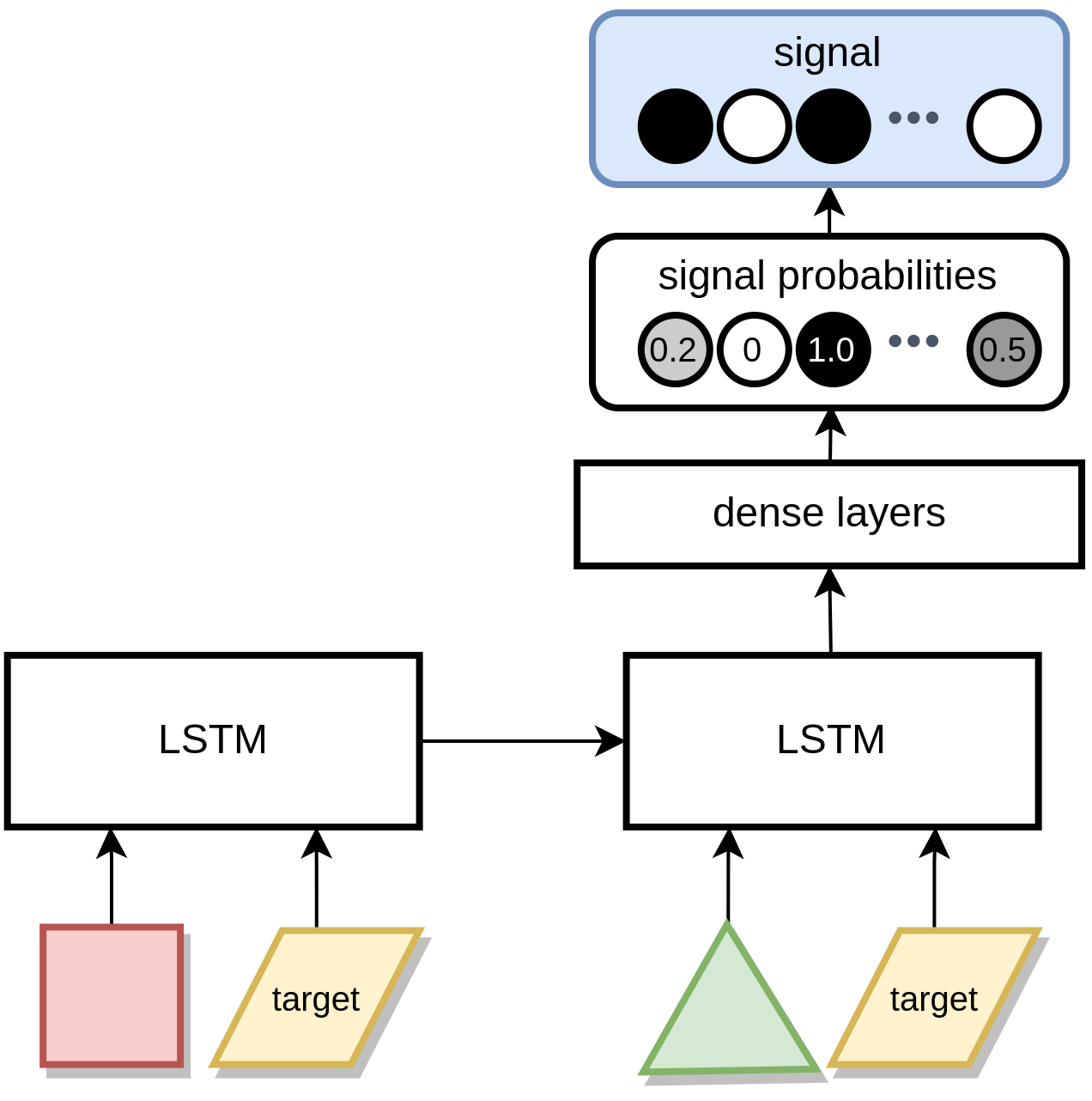}
    \caption{\textbf{Target-in-context (TC) sender}}
    \label{sender_tdp}
\end{figure}

\textbf{Target-in-context (TC) sender }. The input of this network is all objects in the context, including the target. It receives input sequentially (Fig.~\ref{sender_tdp}). As non-target objects are input to the network in subsequent steps, the target is passed along each of them. Such an input structure enables direct comparisons between the target and each of the objects. This network has an LSTM memory cell, which constitutes the first layer of the network. After the last input is provided, LSTM output is passed through fully-connected layers to produce activation probabilities for each symbol. Since this sender architecture is informed by the receiver's context, it is expected to be able to communicate in a context-dependent manner.

\begin{figure}[H]
    \centering
    \includegraphics[scale=0.5]{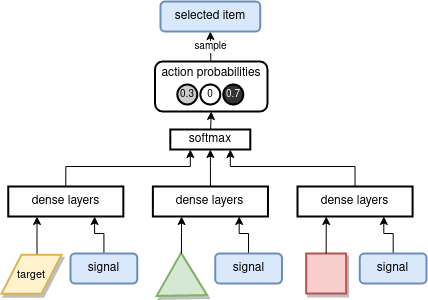}
    \caption{\textbf{The receiver}}
    \label{receiver_mff}
\end{figure}

\textbf{Receiver} The receiver is a simple network that generates the decision on the choice of the object within a single forward pass (Fig.~\ref{receiver_mff}). It is inspired by and very similar to the receiver architecture used by \citet{lazaridou2017} in that each object may only be compared with the received signal independently from other objects. Each object is processed by the same sequence of network layers along with the signal emitted by the sender, which produces what may be interpreted as an alignment score between the input object and the received signal. All such scores are combined with a softmax operation into a probability distribution over objects, from which the final selection is sampled. In simple terms, this receiver most often selects the object that best matches the received signal.

\subsection{Optimization method}

Each iteration in a simulation run consists of $n$ episodes. The fitness of the agents is based on their accuracy and vocabulary size.

\textbf{Accuracy} of a sender-receiver pair characterised by a parameter vector $\theta$ is defined as $Acc(\theta) = \frac{1}{n}\sum_{i=1}^{n} G_i $, where $G_i$ is the reward for the $i$-th episode. $G_i = 1$ if and only if the receiver selects the target object in the $i$-th episode; otherwise $G_i = 0$.

\textbf{Vocabulary size} is defined in a continuous way based on the usage of different signals across all episodes in an iteration. If $\sigma^j$ is the $j$-th symbol (or bit) of the signal vector, $1 \leq j \leq m$, then:

\begin{multline*}
\textit{Voc}(\sigma) = \frac{1}{m}\sum_{j=1}^{m} H(\sigma^j) \\
= -\frac{1}{m}\sum_{j=1}^{m} ( P(\sigma^j = 0)\log_2 P(\sigma^j = 0) \\
+ P(\sigma^j = 1)\log_2 P(\sigma^j = 1) )
\end{multline*}
Intuitively, it means that a language with 10 words that uses only 5 of them in most episodes has a smaller vocabulary than one with 10 words that are equally frequent words. The measure is made continuous for the purpose of convenient gradient optimisation.

\textbf{Fitness} The fitness function calculated for the agent parameters $\theta$ takes the form:
$$   F(\theta) = \frac{1}{n}\sum_{i=1}^{n} G_i - p_{voc} \textit{Voc}(\sigma) $$
where $p_\textit{voc}$ is the penalty coefficient (can be equal to zero).

The model parameters $\theta$ are iteratively optimised to maximise the fitness of the agents. We chose one of the black-box optimisation methods – the evolution strategies \citep{salimans_2017}. We define the model parameters $\theta$ as a concatenation of the sender and receiver parameters, the weights and biases of their corresponding networks. At first, the model parameters are initialised randomly. In every iteration, a mini-batch of new parameters $\theta$ is generated by sampling noise $\epsilon$ from the normal distribution and adding it to the parameter vector. We also use mirror sampling \citep{brockhoff2010mirrored}, that is, for every sampled perturbation $\epsilon$, an inverse perturbation $-\epsilon$ is generated. The perturbed parameters are then evaluated to approximate the gradient of the fitness function $F$:
 $$ \nabla_\theta \mathbb{E}_{\epsilon\sim\mathcal{N}(0, I)} F(\theta + \sigma\epsilon) = \frac{1}{\sigma}\mathbb{E}_{\epsilon\sim\mathcal{N}(0, I)} \left[ \epsilon F(\theta + \sigma\epsilon) \right] $$
The gradient approximated in this way is used for optimisation using gradient ascent. 

\subsection{Measures of communication context-dependence}

To measure how contextual the emerging language is, for each iteration, we calculate several complementary metrics.

\textbf{Target certainty}. This measure is the reverse of the notion of ambiguity, that is, it represents how much each symbol corresponds to only a single target.

At the same time, target certainty is the out-of-context baseline to be compared with actual accuracy. It is the highest accuracy possible in a counterfactual situation such that the sender emits the same signals but the receiver's choice is neither constrained nor informed by context. Therefore, the best strategy is to select the most probable target given the given signal. We expect that the more the language becomes context-dependent, the more ``imperfect'' it will be in the eyes of an observer that considers only the received signal.

$$ \frac{1}{n}\sum_{i = 1}^n \max_{\textit{o} \in \textit{O}} P(T = \textit{o} | \sigma_i) $$

\textbf{Signal certainty}. This is an analogue of target certainty on the sender's end -- a measure of how much each target perceived by the sender maps to a single signal.

$$ \frac{1}{n}\sum_{i = 1}^n \max_{\sigma \in \Sigma} P(\sigma | t_i) $$

\textbf{Maximal contextless interpretation accuracy}. This measure is devised to detect nontrivial context processing by the receiver. It corresponds to accuracy in a counterfactual situation where the receiver is neither constrained nor informed by the context and still ``does its best''. In this scenario, the sender’s signal emission is unchanged, but the receiver selects the most probable target in the current context given the received signal alone. If the real accuracy exceeds this hypothetical quantity, it means that the receiver conditions its response on the context in a more complex way than choosing the most frequent signal referent considering each object $o$ present in the context.

$$ \frac{1}{n}\sum_{i = 1}^n 1 \textit{ iff } t_i = \arg \max_{\textit{o} \in c_i}  P(T = \textit{o} | \sigma_i) \textit{ else } 0 $$

For a full picture of the emergent communication, we additionally calculate information-theoretic measures based on Shannon's mutual information \citep{cover2012}, here denoted by $I$. They quantify context's participation in the signal generation on the part of the sender and in the interpretation of its meaning on the part of the receiver – which here amounts to the selection of the object from the context. For brevity, exact values of these metrics are reported in Appendix \ref{appendix}.

\textbf{Sender context information gain}.

This is the amount of information gained about the signal when the whole context is considered beyond the target. When $C$, $T$ and $\sigma$ correspond to random variables denoting context, target, and signal emitted, we define the metric as follows:
$ I(\sigma;C) - I(\sigma;T) $

\textbf{Receiver context information gain}.

This is the amount of information gained about the receiver response when the context is considered beyond the received signal. When $R$ is the receiver's response random variable, this metric is defined as follows:
$ I(R;C,\sigma) - I(R;\sigma) $


\section{Simulation results}

We ran our model in the world where all objects have $2$ categorical properties, both with $3$ possible values. For illustration purposes, we can consider them as shape and colour. The context size was set to $3$. \added{The context was entirely random, that is, each object appeared in the context with equal probability and, consequently, so did every possible object combination. The target, too, was selected from the context at random.}

We tested our models in two main conditions: with unlimited vocabulary and with a vocabulary size penalty. The experiments were carried out with both T and TC~senders (see section \ref{sec:contextual_model}), paired with the same receiver type. We ran each simulation $10$ times for $50k$ iterations. The exact model hyperparameters used in both experiments and final metric values are specified in Appendix \ref{appendix}.

 \begin{figure*}
    \centering
    \begin{tabular}{cc}
        (a) \quad \raisebox{1ex-\height}{\includegraphics[scale=0.77]{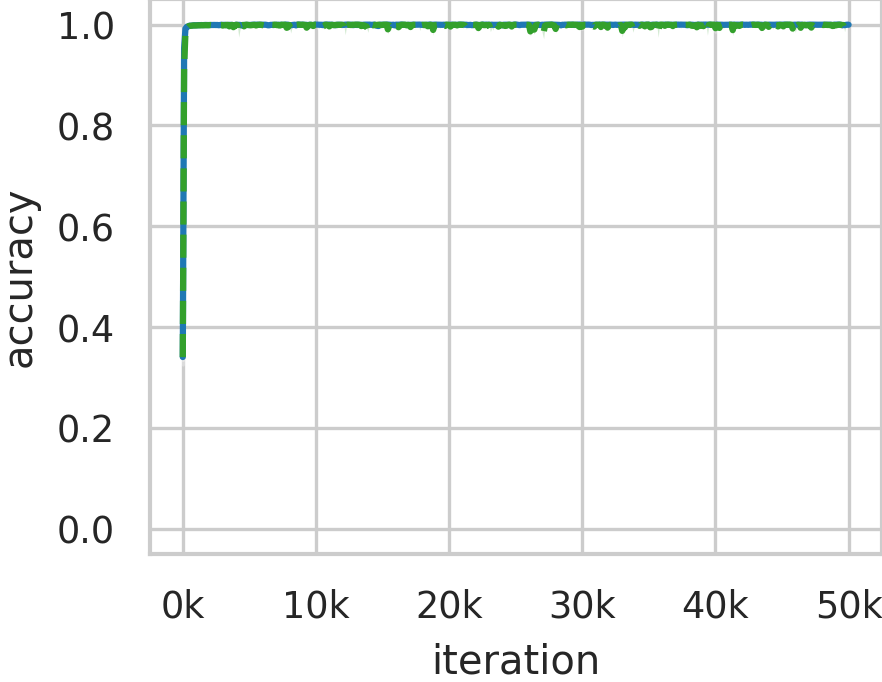}} 
        & 
        (b) \quad \raisebox{1ex-\height}{\includegraphics[scale=0.77]{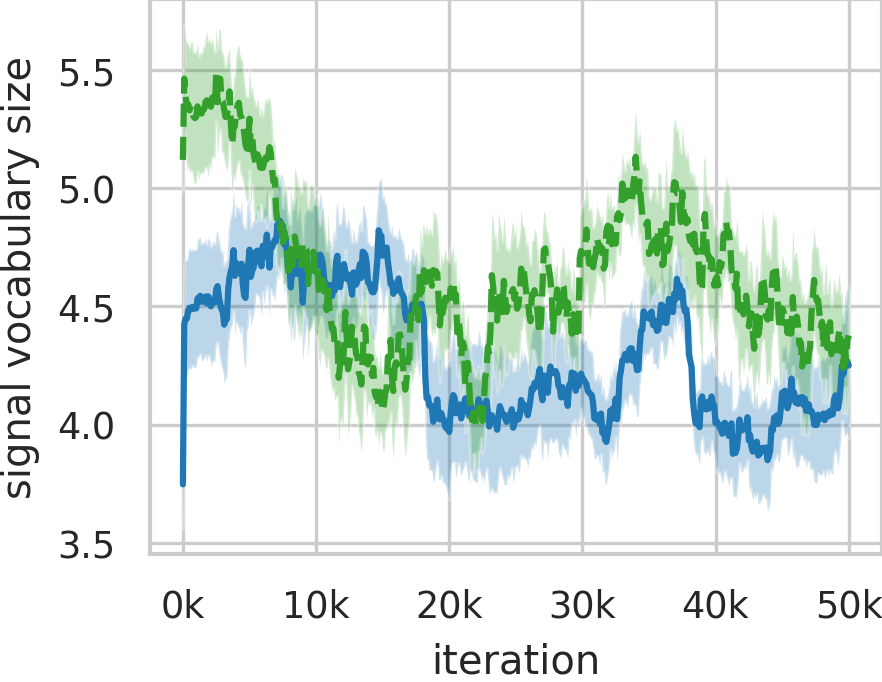}}
        \\[0.2cm]
        \rule{0pt}{4ex}   
        (c) \quad \raisebox{1ex-\height}{\includegraphics[scale=0.77]{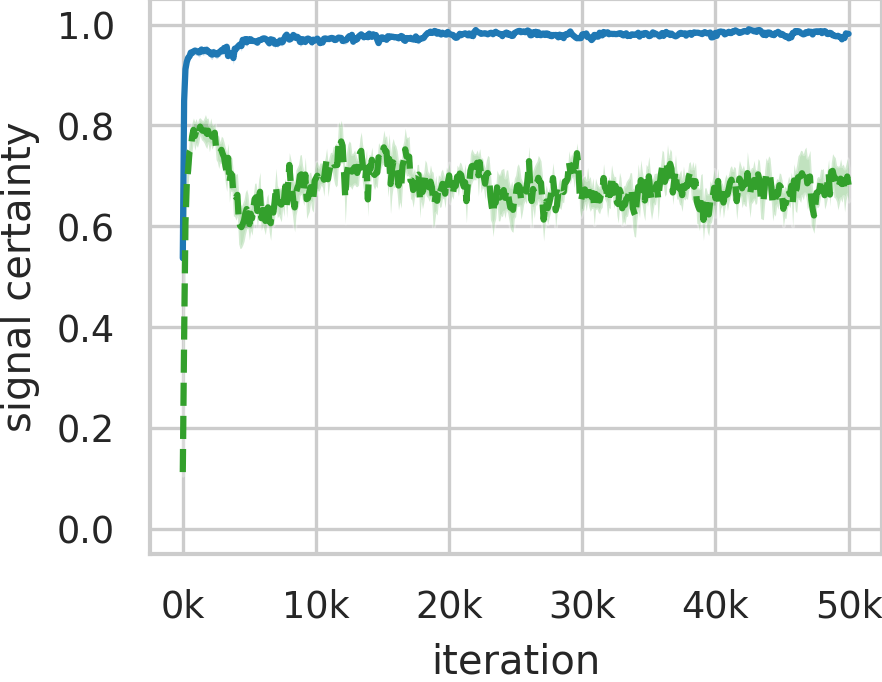}}
        &
        (d) \quad \raisebox{1ex-\height}{\includegraphics[scale=0.77]{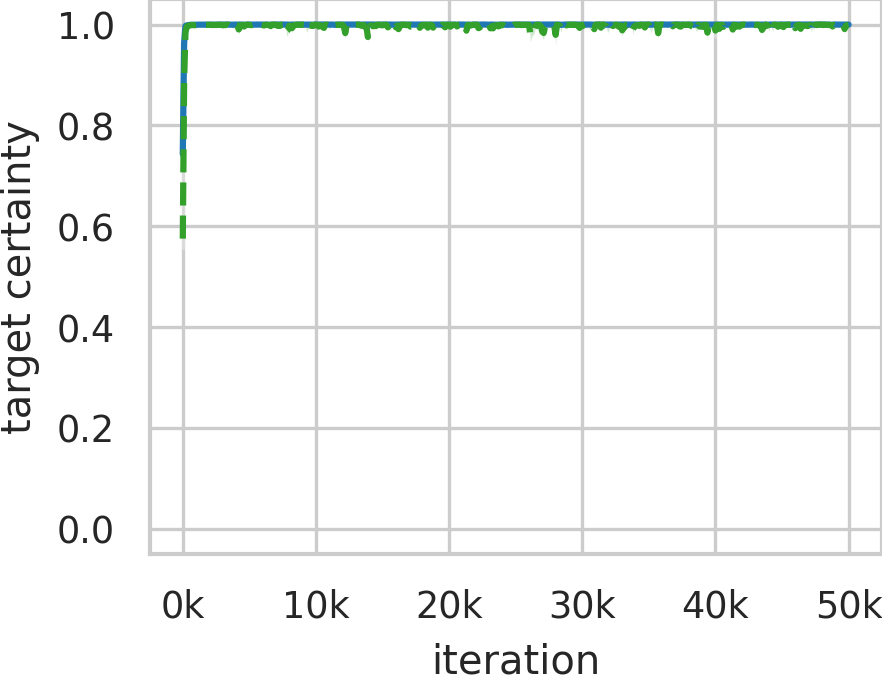}}
        \\[0.2cm]
        \multicolumn{2}{c}{
            \rule{0pt}{4ex}   
            \includegraphics[scale=0.77]{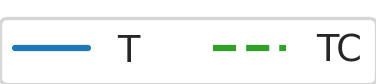}
        } \\[0.2cm]
    \end{tabular}
    \caption{Metric plots for Experiment 1: T and TC~sender with no vocabulary penalty. The dark area around curves marks the standard error.}
    \label{fig:exp1_plot_results}
\end{figure*}

\begin{figure*}
    \centering
    \begin{tabular}{cc}
        (a) \quad \raisebox{1ex-\height}{\includegraphics[scale=0.77]{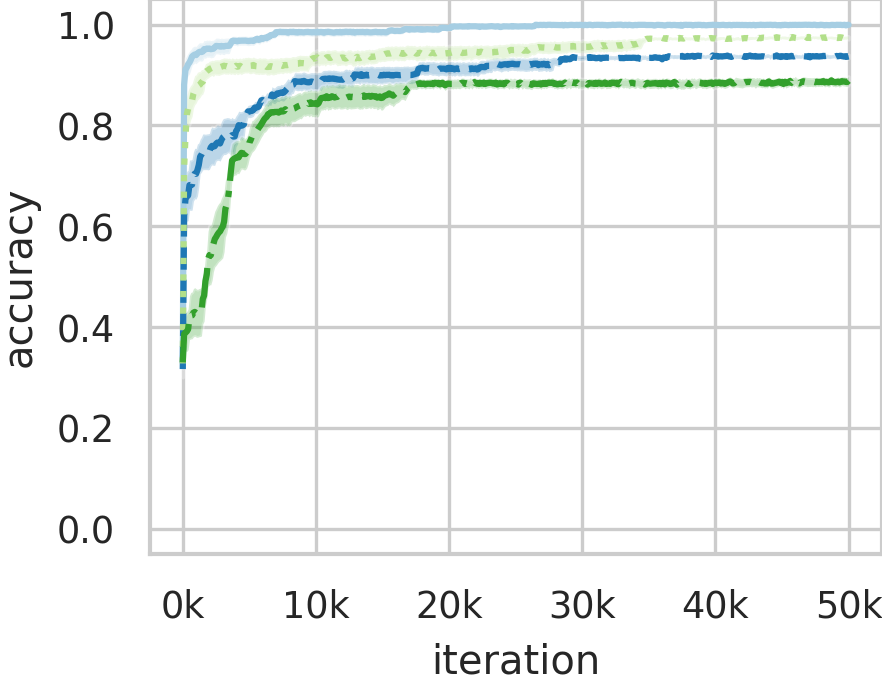}}
        &
        (b) \quad \raisebox{1ex-\height}{\includegraphics[scale=0.77]{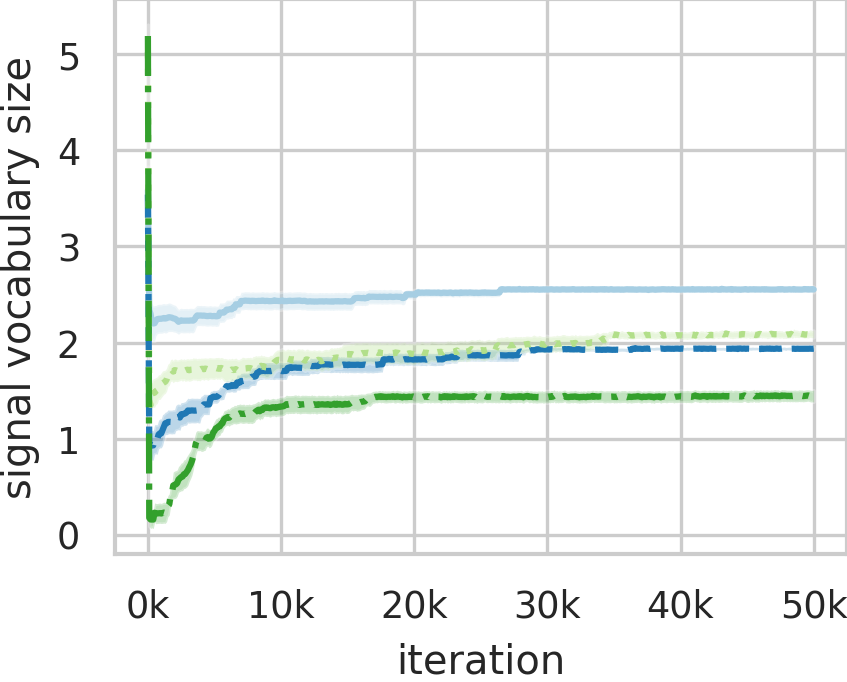}}
        \\[0.2cm]

        \rule{0pt}{4ex} 
        (c) \quad \raisebox{1ex-\height}{\includegraphics[scale=0.77]{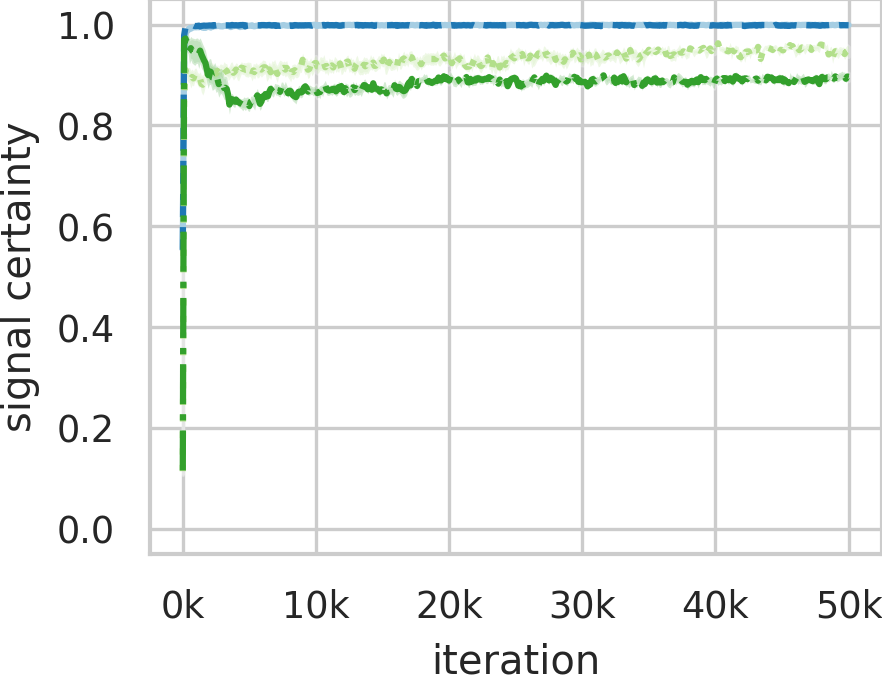}}
        & 
        (d) \quad \raisebox{1ex-\height}{\includegraphics[scale=0.77]{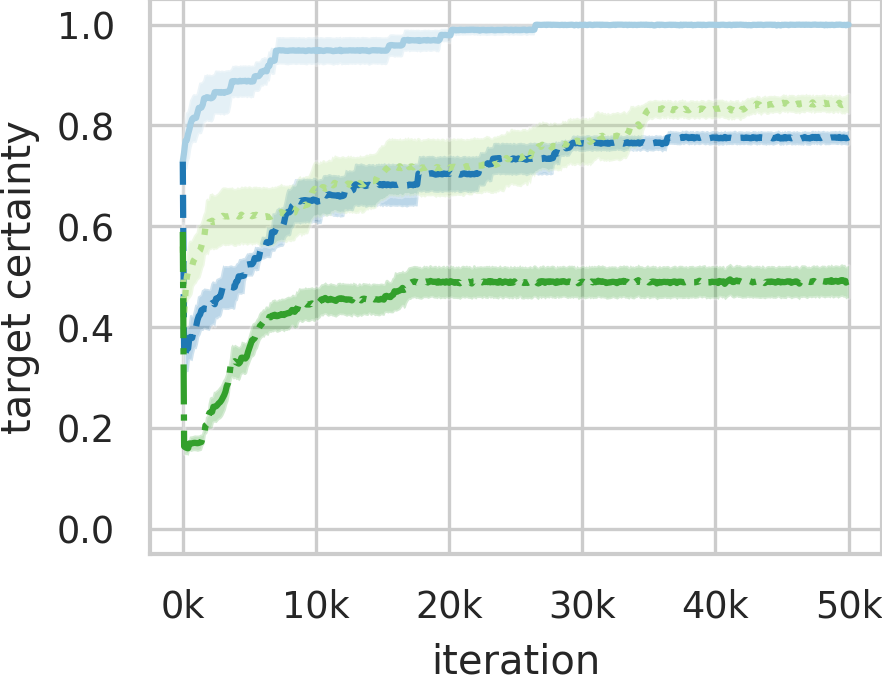}}
        \\[0.2cm]
        
        \multicolumn{2}{c}{
            \rule{0pt}{4ex} \includegraphics[scale=0.77]{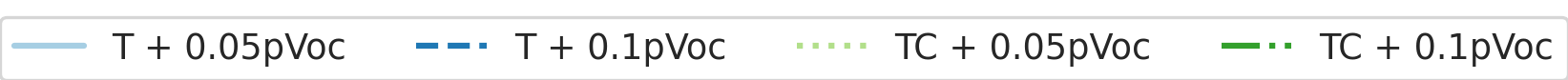} 
        } 
        \\
    \end{tabular}
    \caption{Metric plots for Experiment 2: T and TC~sender combined with 2 different levels of vocabulary penalty. The dark area around curves marks the standard error.}
    \label{fig:exp2_plot_results}
\end{figure*}

\subsection{Experiment 1: Unlimited vocabulary}
\label{sec:experiment1}

The goal of the first experiment was to check whether agents could develop contextual communication without additional pressure beyond the pressure for communication accuracy. The results can tell us whether the agent architectures used in this study (a) have any inherent tendency to develop contextual languages and (b) can reach maximum accuracy in a favourable setting. In this experiment, the vocabulary penalty coefficient $p_\textit{voc} = 0$.

Regardless of the sender type, whether T or TC, all agent pairs converge to top accuracy (Fig.~\ref{fig:exp1_plot_results}a). The developed languages are fully unambiguous, as shown by the maximum value of target certainty -- given the signal, the receiver selects the correct object unequivocally regardless of the remaining context. Interestingly, TC~sender, unlike T~sender, does not always send the same signal given the target (Fig.~\ref{fig:exp1_plot_results}c). However, this does not amount to context-dependent communication, since at the receiver's end the signal always unequivocally indicates the target, which is shown by the maximum value of target certainty (Fig.~\ref{fig:exp1_plot_results}d). Let us note that, nevertheless, receiver context information gain is non-zero, nearing $1.2$ for both receiver types: the context is informative about the receiver's response as it constrains it directly.

\subsection{Experiment 2: Penalised vocabulary size}
\label{sec:experiment2}

In this experiment, the fitness of the agents is reduced proportionally to the size of their vocabulary. According to previous research \citep{santana2014ambiguity, muhlenbernd2021evolutionary}, this should introduce pressure for contextual communication. We expected that TC~sender's access to context should allow it to ``get away'' with ambiguity, which would lead to languages with a smaller vocabulary. T~sender, since it does not perceive the receiver’s context, should not be able to communicate in a context-dependent manner despite the pressure that favours context-dependence. We tested two different levels of the vocabulary penalty coefficient $p_\textit{voc}$: $0.05$ and $0.1$.

\begin{figure}[H]
    \centering
    \begin{tabular}{c}
        \includegraphics[scale=0.5]{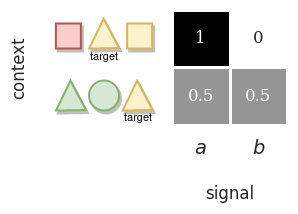} \\
    \end{tabular}
    \caption{
    \replaced{Signal emission probabilities for selected target and contexts in an exemplary run of \textit{TC + 0.1pVoc} condition}{Signal emission probabilities for \textit{TC + 0.1pVoc} condition for an exemplary target and exemplary contexts}. Given yellow triangle target, the sender always sends one of two signals. Our inspection of the resulting protocol shows that depending on the context, the sender's signal follows one of two distinct distributions, exemplified here. 1) For most contexts, signal $a$ is sent with maximum probability, but 2) if green triangle also appears in the context, one of the two signals, either $a$ or $b$ is sent with near-equal probability.
    }
    \label{fig:audience_design_example}
\end{figure}

\begin{figure}[H]
    \centering
    \begin{tabular}{c}
        \includegraphics[scale=0.5]{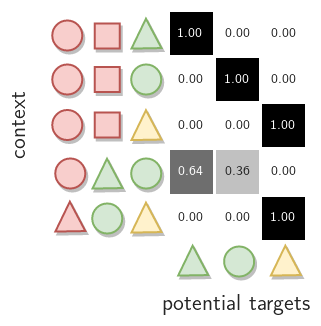} \\
    \end{tabular}
    \caption{
    \replaced{Meaning probabilities for selected signal and contexts in an exemplary run of \textit{TC + 0.1pVoc} condition.
    }{Meaning probabilities for \textit{TC + 0.1pVoc} condition for exemplary signal $a$ and exemplary contexts.} \replaced{This signal}{Signal $a$}, across all contexts, always refers to one of three objects. The bottom two contexts show two different cases of how potentially conflicting meanings are resolved. When both green triangle and green circle are present, the signal is dysfunctionally ambiguous as it may refer to either of them. When both green circle and yellow triangle are present, there is one conventionalized referent: yellow triangle.
    }
    \label{fig:receiver_language_example}
\end{figure}

All emergent signal systems were influenced by pressure to minimise their vocabulary size, which now decreased from values slightly above $4$ in Experiment 1 to values ranging from $1.4$ to $2.6$ (Fig.~\ref{fig:exp2_plot_results}b). Significant signal ambiguity, indicated by reduced target certainty (Fig.~\ref{fig:exp2_plot_results}d), appeared in all conditions except \textit{T + 0.05pVoc}.\deleted{Figure ~\ref{fig:audience_design_example} qualitatively describes the strategy used by TC~sender, while Figure ~\ref{fig:receiver_language_example} additionally shows some of the language-meaning links from the receiver's perspective.}

For all agents with ambiguous signals, the ambiguity came at the price of slightly reduced accuracy (Fig.~\ref{fig:exp2_plot_results}a). Nevertheless, the accuracy was still relatively high (above $0.88$) compared to the target certainty ($0.49$ to $0.78$) (Fig.~\ref{fig:exp2_plot_results}d). This means that the signal alone without contextual information and constrained object choice would only allow for much lower accuracy. This effect could be observed to some degree for both sender types. The achieved accuracy did not exceed the maximum contextless interpretation accuracy for either type of sender (see Appendix \ref{appendix} for exact values), confirming that the receiver architecture did not allow for any sort of contextual inference.

There are differences between T and TC~senders. At near-identical fitness levels, TC~sender managed to reduce its vocabulary in comparison to T~sender. While T~sender could only produce symbols based on the target, TC~sender's signal certainty values are below maximum (Fig.~\ref{fig:exp2_plot_results}c) as the emitted signal can be conditional on the context. The target certainty was much lower for TC than for T~sender for both penalty levels. In other words, TC~sender managed to ``reuse'' the same signals to a greater degree than T~sender, which resulted in greater ambiguity of the signals actually used.

\begin{figure}[H]
    \centering
    \begin{tabular}{c}
        \includegraphics[scale=0.55]{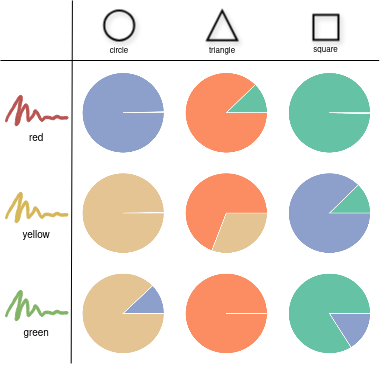} \\
        \includegraphics[scale=1]{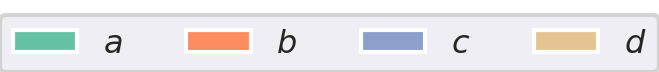} \\
    \end{tabular}
    \caption{
        \added{
        Signal emission frequencies for an exemplary run of \textit{TC + 0.1pVoc} condition, mapped for all $9$ target objects. Signal names in the legend are arbitrary. $b$ refers to all triangles, so it is the most systematic of all $4$ signals. On the other end of this spectrum, signal $c$ is used to refer to many unrelated meanings: red circles, yellow squares, but also sometimes to green circles and green squares.
        }
    }
    \label{fig:mapping_comp_example}
\end{figure}

\added{
We examined in more detail languages used in the final iteration of \textit{TC + 0.1pVoc} condition. In $9$ out of $10$ runs, $4$ most frequent signals amounted to more than $99\%$ of all emitted signals. In the remaining one, the same percentage was comprised by $7$ most frequent signals.
For all runs, the sender's signal use formed diverse patterns across all $9$ communicated targets (Fig.~\ref{fig:mapping_comp_example}): including signals systematically used for one category (e.g. triangles of all colours), signals used for disjoint meanings (e.g. red circle and yellow square), as well as combinations of both. While some targets were always communicated using the same signal, for others the sender's signal choice depended on other items in the context (Fig.~\ref{fig:audience_design_example}).
The receiver's choice, too, was context-dependent for some signals (Fig.~\ref{fig:receiver_language_example}): deterministic and correct when only one likely target was present but unreliable under meaning conflict.
}

\section{Discussion}
\label{sec:discussion}

We investigated a situated communicative scenario in which the environment directly constrains the receiver's response to the objects in the current context and thus constrains its interpretation of the signal. In this setting, we examined the consequences of pressure to minimise vocabulary. Our results are consistent with previous modelling work in that the penalty on vocabulary size induced \replaced{a degree of}{out-of-context language} ambiguity \citep{santana2014ambiguity, o2015ambiguity, muhlenbernd2021evolutionary}. Furthermore, they show the ambiguity that arises as a trade-off between accuracy and minimisation of vocabulary size, in accordance with \citet{o2015ambiguity} predictions that a degree of inaccuracy is optimal for a certain ratio between the cost of vocabulary size and the size of the meaning space to be expressed. In Experiment 2, with a large enough vocabulary size penalty, the accuracy dropped below the maximum value for both sender types. 

In our model, unlike in the above works and others  \citep{piantadosi2012}, effective disambiguation of signal meaning depends on the environmental constraints on the referent choice, and not on the receiver's inferential capabilities. The effectiveness of disambiguation can be seen in that the accuracy reached levels higher than {our out-of-context baseline -- target certainty -- for both types of senders. While the target-in-context sender (TC) did not outperform the target-only sender (T) in terms of raw accuracy, it developed a more ambiguous language with a smaller vocabulary that was still reasonably effective. This was the case even though the receiver was always context-agnostic. The context-aware sender was able to capitalise on the situated context constraining receiver's actions. This kind of context-dependent signalling constitutes a variant of audience design that does not rely on simulating inferential processes at the receiver's end, as in some previous works \citep{batali1998, brochhagen2020signalling}. The need for context awareness on the sender's side is consistent with empirical experiments by \citet{winters2018contextual}, where the most sophisticated languages that encoded only the contextually relevant semantic distinctions emerged only when the sender was able to perceive the context.

Disambiguation mechanisms similar to those present in our model arguably play a major role in shaping languages; natural languages evolve under pressures that favour ambiguity, such as minimising speaker effort \citep{piantadosi2012} or language \replaced{learnability}{simplicity} \citep{kirby_2015}, and these mechanisms may allow ambiguity to enter the language without sacrificing communication success. Although constrained referent choice in the absence of the sender's context awareness does not offer a systematic disambiguation mechanism, it may still greatly reduce communication errors and thus pressure for meaning precision. As the two meanings of the word \textit{bat} -- the baseball bat and the flying mammal -- are unlikely to be relevant in the same context, the word itself may safely remain ambiguous. When the sender becomes aware of the context, different or additional linguistic means can be used to specify the meaning. This was illustrated by the context-aware sender's behaviour in our model and reflected in an experiment by \citet{ferreira2005speakers}, where participants were tasked to describe a scene. Perceiving both the flying mammal and a baseball bat in one scene, the participants often used an additional specifier for one of the objects, producing descriptions such as ``bat and baseball bat''. Clearly, complex human inferential capabilities that involve both sender and receiver, as considered in pragmatics (e.g. \citet{korta2020}), will likely make the ambiguous language even more effective.

While context-dependent communication has usually been portrayed as relying on processes internal to the communicating agents, we demonstrate that a nontrivial part of the job of disambiguation is performed by the environment itself \citep{brooks1990elephants, wilson2013embodied}. Consequently, it seems that the composition of contexts in which we find ourselves plays a key role. \citet{guo2021expressivity} show in a computational model that languages developed by neural agents in signalling games with medium context size encode more information than those emerging within a smaller or larger context. We can speculate that another source of environmentally induced ambiguity may be simply the frequency of referents. \citet{huttegger2007evolution} shows that for an uneven distribution of referent types, the population is more likely to stay in a suboptimal ambiguous signalling strategy. As some types need to be communicated less frequently, the potential payoff of assigning them a unique symbol is reduced. In our simulations, we used a very simple environment with uniform distribution over the communicated targets and uncorrelated object occurrence within the context. More complex environments with predictable regularities open new possibilities. If two particular objects never co-occur (the baseball bat and the flying mammal), the same signal can be potentially used for both of them without causing communication errors. \citet{winters2018contextual} show that predictable composition of the context is, in fact, exploited by human participants who, consequently, arrive at simpler languages than those emerging for entirely random contexts. 

There may be more mechanisms responsible for handling ambiguity in language than is often assumed. The phenomena discussed so far concerned one-directional communication. In contrast, human interlocutors engage in dialogues where they explicitly ask for disambiguation \citep{healey2018} or even co-construct meaning together \citep{kempson2019}. Looking from this angle, we discover that not only sharing context, but also aggregating information from individually available contexts is important for dialogically co-constructed messages \citep{raczaszek2014pooling}. These multiple mechanisms can remedy meaning uncertainty when precision is vital; the very same uncertainty can allow for language flexibility, which is indispensable on other occasions (see Section~\ref{sec:ling_context}).

There are a couple of directions where the current work can be developed. First, future research could examine the effects of various context properties, such as the distribution of objects in the environment. In our work, we used only a uniform distribution; moving away from it would affect relative benefits of distinguishing certain object properties and also introduce regularities that could be exploited by agents in their disambiguation strategies.
Second, the analysis could be extended to answer questions concerning emerging language structure, an issue that we have left aside in our work. As in natural languages a neighbouring word often sufficiently disambiguates an otherwise ambiguous expression \citep{ferreira2008ambiguity}, we could expect that complexifying signals would render their individual components even more ambiguous. One possible amendment to the model would be allowing agents to emit compound signals of variable length.
Finally, the current model could be altered to allow bi-directional exchange between agents (in the vein of \citet{kottur_2017, bancerek2023}). This would allow us to investigate dialogical mechanisms of disambiguation and their interactions with situational context in constructing flexible language.

\section{Conclusion}
\label{sec:conclusion}

In this work, we developed a model in which the pressure that is conducive to the emergence of context-dependent communication was combined with the environmentally constrained interpretation of the signal by the receiver. In doing this, we complemented previous modelling work, which focused on how communicating agents can exploit shared information by means of inference. To the best of our knowledge, this is also the first work where environmentally constrained communication was modelled in conjunction with a vocabulary size cost.
Our results point to the fact that context-dependent communication is likely a multilayered phenomenon, which should be analysed not only in terms of individual inferential capabilities, but also in terms of the agents-environment \added{and agent-agent} relations. In line with other authors of recent research on linguistic context dependence, we urge that out-of-context signal ambiguity should not be readily equated with proneness to error but rather understood as a pervasive property of real-life situated communication, resulting in communicative efficiency and flexibility.

\section*{Acknowledgements}
This work was funded by the National Science Centre, Poland (OPUS 15 grant, 2018/29/B/HS1/00884). The experiments were partly run on the supercomputers of the Interdisciplinary Centre for Mathematical and Computational Modelling at the University of Warsaw (computational grant G86-1039). The metrics in the experimental results were tracked via Weights \& Biases \citep{biewald_2020}.

\section*{Declarations of interest}
The authors declare that they have no competing interests.

\bibliography{article} 

}

\end{multicols}

\pagebreak

\appendix
\section{Appendix}
\label{appendix}

\subsection{Model hyperparameters}
This section lists the details of network architecture and hyperparameters used for both experiments in this work.
All non-LTSM network layers are interleaved with ReLU activations. TC~sender consists of a single LSTM layer and $2$ linear layers. T~sender and the receiver network are both made of $3$ linear layers.
The hyperparameter values are manually selected to allow smooth optimisation. Hyperparameters below describe all networks in this work:

Network structure:
\begin{itemize}
	\item The size of all dense layers: $50$
	\item The size of the hidden state and the cell state for LSTM cells: $50$
\end{itemize}

Optimisation with evolution strategies
\begin{itemize}
	\item mini-batch size: $50$
	\item episodes per iteration: $400$
	\item sample noise standard deviation $\sigma$: $0.1$
	\item gradient optimiser: AdamW with weight decay $0.01$ and learning rate dependent on the sender type:
	\begin{itemize}
        \item $0.02$ for the receiver
	    \item $0.05$ for TC~sender and $0.02$ for T~sender

    \end{itemize}
\end{itemize}

The results of \citet{lazaridou_2018} suggest that a too low signal length may make the model more prone to getting stuck in the local minima; we set a relatively large signal length $10$, to ensure that such a problem does not occur.

\subsection{Experiment results}

\begin{table}[htb]
	\caption{Final metric values for Experiment 1}
	\centering
	\begin{tabular}{l|l|r|r|r|r}
\toprule
run & metric & mean & std & min & max \\
\midrule
T & fitness & 1.000 & 0.001 & 0.998 & 1.000 \\
TC & fitness & 0.999 & 0.001 & 0.996 & 1.000 \\
T & accuracy & 1.000 & 0.001 & 0.998 & 1.000 \\
TC & accuracy & 0.999 & 0.001 & 0.996 & 1.000 \\
T & signal vocabulary size & 4.250 & 0.961 & 3.181 & 5.605 \\
TC & signal vocabulary size & 4.378 & 0.648 & 3.372 & 5.048 \\
T & receiver context information gain & 1.220 & 0.005 & 1.211 & 1.226 \\
TC & receiver context information gain & 1.218 & 0.005 & 1.212 & 1.228 \\
T & max contextless interpretation accuracy & 1.000 & 0.000 & 1.000 & 1.000 \\
TC & max contextless interpretation accuracy & 1.000 & 0.000 & 0.999 & 1.000 \\
T & sender context information gain & 0.035 & 0.027 & 0.002 & 0.081 \\
TC & sender context information gain & 0.780 & 0.274 & 0.250 & 1.128 \\
T & signal entropy & 3.217 & 0.048 & 3.158 & 3.294 \\
TC & signal entropy & 4.231 & 0.368 & 3.504 & 4.732 \\
T & signal certainty & 0.982 & 0.014 & 0.962 & 0.999 \\
TC & signal certainty & 0.682 & 0.104 & 0.563 & 0.896 \\
T & target certainty & 1.000 & 0.000 & 0.999 & 1.000 \\
TC & target certainty & 0.999 & 0.001 & 0.996 & 1.000 \\
\bottomrule
\end{tabular}

	\label{tab:exp1_full_results}
\end{table}

\begin{table}[htb]
	\caption{Final metric values for Experiment 2}
	\centering
	\begin{tabular}{l|l|r|r|r|r}
\toprule
run & metric & mean & std & min & max \\
\midrule
T + 0.05pVoc & fitness & 0.872 & 0.005 & 0.865 & 0.878 \\
T + 0.1pVoc & fitness & 0.743 & 0.007 & 0.729 & 0.748 \\
TC + 0.05pVoc & fitness & 0.869 & 0.007 & 0.859 & 0.880 \\
TC + 0.1pVoc & fitness & 0.744 & 0.010 & 0.730 & 0.765 \\
T + 0.05pVoc & accuracy & 0.999 & 0.001 & 0.998 & 1.000 \\
T + 0.1pVoc & accuracy & 0.937 & 0.012 & 0.915 & 0.944 \\
TC + 0.05pVoc & accuracy & 0.973 & 0.010 & 0.960 & 0.983 \\
TC + 0.1pVoc & accuracy & 0.889 & 0.027 & 0.863 & 0.963 \\
T + 0.05pVoc & signal vocabulary size & 2.552 & 0.095 & 2.406 & 2.685 \\
T + 0.1pVoc & signal vocabulary size & 1.937 & 0.044 & 1.853 & 1.967 \\
TC + 0.05pVoc & signal vocabulary size & 2.089 & 0.113 & 1.990 & 2.297 \\
TC + 0.1pVoc & signal vocabulary size & 1.444 & 0.191 & 1.331 & 1.982 \\
T + 0.05pVoc & receiver context information gain & 1.218 & 0.006 & 1.207 & 1.227 \\
T + 0.1pVoc & receiver context information gain & 1.129 & 0.018 & 1.092 & 1.146 \\
TC + 0.05pVoc & receiver context information gain & 1.196 & 0.010 & 1.178 & 1.209 \\
TC + 0.1pVoc & receiver context information gain & 1.108 & 0.026 & 1.084 & 1.179 \\
T + 0.05pVoc & max contextless interpretation accuracy & 1.000 & 0.000 & 1.000 & 1.000 \\
T + 0.1pVoc & max contextless interpretation accuracy & 0.944 & 0.010 & 0.922 & 0.950 \\
TC + 0.05pVoc & max contextless interpretation accuracy & 0.975 & 0.010 & 0.962 & 0.984 \\
T + 0.05pVoc & sender context information gain & 0.002 & 0.001 & 0.000 & 0.004 \\
T + 0.1pVoc & sender context information gain & 0.003 & 0.003 & 0.001 & 0.011 \\
TC + 0.05pVoc & sender context information gain & 0.123 & 0.060 & 0.052 & 0.208 \\
TC + 0.1pVoc & sender context information gain & 0.259 & 0.052 & 0.126 & 0.305 \\
T + 0.05pVoc & signal entropy & 3.157 & 0.003 & 3.154 & 3.162 \\
T + 0.1pVoc & signal entropy & 2.673 & 0.093 & 2.494 & 2.727 \\
TC + 0.05pVoc & signal entropy & 2.913 & 0.083 & 2.787 & 3.036 \\
TC + 0.1pVoc & signal entropy & 2.065 & 0.247 & 1.908 & 2.764 \\
T + 0.05pVoc & signal certainty & 0.999 & 0.001 & 0.998 & 1.000 \\
T + 0.1pVoc & signal certainty & 0.999 & 0.002 & 0.994 & 1.000 \\
TC + 0.05pVoc & signal certainty & 0.946 & 0.032 & 0.899 & 0.983 \\
TC + 0.1pVoc & signal certainty & 0.896 & 0.026 & 0.855 & 0.955 \\
T + 0.05pVoc & target certainty & 1.000 & 0.001 & 0.999 & 1.000 \\
T + 0.1pVoc & target certainty & 0.776 & 0.042 & 0.696 & 0.800 \\
TC + 0.05pVoc & target certainty & 0.843 & 0.060 & 0.769 & 0.892 \\
TC + 0.1pVoc & target certainty & 0.490 & 0.101 & 0.450 & 0.778 \\
\bottomrule
\end{tabular}
	\label{tab:exp2_full_results}
\end{table}


\end{document}